\documentclass[runningheads]{llncs}
\usepackage[T1]{fontenc}
\usepackage{graphicx}

\usepackage[export]{adjustbox}
\usepackage[colorlinks=true, allcolors=blue]{hyperref}
\usepackage{multirow}
\usepackage{booktabs}
\usepackage{rotating}
\usepackage{subcaption}
\usepackage{amssymb}

\begin{document}
\title{A Three-Way Knot: Privacy, Fairness, and Predictive Performance Dynamics}
%
%

\author{Tânia Carvalho\inst{1}\orcidID{0000-0002-7700-1955} \and
Nuno Moniz\inst{2}\orcidID{0000-0003-4322-1076} \and
Luís Antunes\inst{1,3}\orcidID{0000-0002-9988-594X}}
\authorrunning{Carvalho et al.}
%
\institute{Faculty of Computer Science, University of Porto, Portugal \and
Lucy Family Institute for Data \& Society, University of Notre Dame, Indiana, USA \and TekPrivacy, Porto, Portugal
\email{tania.carvalho@fc.up.pt}}
%

\maketitle              
\begin{abstract}
As the frontier of machine learning applications moves further into human interaction, multiple concerns arise regarding automated decision-making. Two of the most critical issues are fairness and data privacy. On the one hand, one must guarantee that automated decisions are not biased against certain groups, especially those unprotected or marginalized. On the other hand, one must ensure that the use of personal information fully abides by privacy regulations and that user identities are kept safe. The balance between privacy, fairness, and predictive performance is complex. However, despite their potential societal impact, we still demonstrate a poor understanding of the dynamics between these optimization vectors. In this paper, we study this three-way tension and how the optimization of each vector impacts others, aiming to inform the future development of safe applications. In light of claims that predictive performance and fairness can be jointly optimized, we find this is only possible at the expense of data privacy. Overall, experimental results show that one of the vectors will be penalized regardless of which of the three we optimize. Nonetheless, we find promising avenues for future work in joint optimization solutions, where smaller trade-offs are observed between the three vectors.

\keywords{Synthetic Data \and Privacy \and Fairness \and Predictive Performance}
\end{abstract}
\section{Introduction}\label{sec:intro}

Growing privacy concerns have led to several approaches aiming to preserve the confidentiality of individuals' information. Among the most prevalent approaches to privacy preservation is the process of data synthesis, which mimics the original data while maintaining its global properties~\cite{carvalho2023survey}. The creation of synthetic data offers a promising avenue, as it generates a protected version of the original data that can be publicly available. Usually, approaches for data synthetization include sampling methods or deep learning-based models~\cite{figueira2022survey}. However, despite significant progress in recent years, the challenge of synthetic data generation methods in preserving the confidentiality of personal data and generating unbiased and accurate machine learning models remains an ongoing area of research and development. The interplay between privacy, fairness, and predictive performance in synthetic data generation is a fundamental issue that requires attention to facilitate the responsible utilization of data in machine learning applications. 

This paper explores the dynamics between preserving privacy and improving fairness and predictive performance in machine learning models. First, to address privacy concerns, we apply privacy-preserving techniques for secure data publication, particularly data synthetization methods, where each synthetic data variant is evaluated concerning its re-identification risk. Then, in the evaluation process for fairness and predictive performance, models are trained and optimized for each synthetic data variant using fairness-agnostic (standard machine learning algorithms) and fairness-aware algorithms. Our main goal is to discover the dynamics of optimizing each vector. The experiments conducted in this study use some of the most popular data sets in FAccT\footnote{Acronym for Fairness, Accountability, and Transparency.} research~\cite{le2022survey,chakraborty2021bias}.

The main conclusions of this work indicate that \textit{1)} solutions that achieve a balance between predictive performance and fairness are only possible at the expense of data privacy, and \textit{2)} generally, optimizing any of the vectors will impact at least another one, but \textit{3)} three-way optimization demonstrates promise for future research.

The remainder of the paper is organized as follows. Section~\ref{sec:background} includes some preliminaries on privacy and fairness in machine learning and overviews of related work on this topic. The experimental study is described in Section~\ref{sec:experiments}, including a description of data, methods, and results. Section~\ref{sec:discussion} discusses such results. Conclusions are provided in Section~\ref{sec:conclusion}.

\section{Background}\label{sec:background}

Existing literature outlines the key concepts of identifying and measuring privacy and algorithmic fairness in machine learning applications~\cite{caton2020fairness,mehrabi2021survey}. In the subsequent sections, we provide concise definitions of relevant background knowledge.

\subsection{Privacy}

Releasing or sharing data about individuals often implies de-identifying personal information for privacy preservation~\cite{carvalho2023survey,torra2022guide}. Conventional de-identification approaches involve applying privacy-preserving techniques such as generalization or suppression to reduce data granularity or introduce noise to the data causing distortion. These transformations are usually applied to a set of quasi-identifiers, i.e., attributes that, when combined, generate a unique signature that may lead to re-identification (e.g., date of birth, gender, profession, and ethnic group), as well as sensitive attributes like religion and sexual orientation which are highly critical. In the case of synthetic data generation, de-identification is generally performed for all attributes and instances to capture the overall characteristics, creating a new data set through generative models~\cite{figueira2022survey}.

Even in a de-identified data set, it is crucial to evaluate the privacy risks as it is challenging to know who the intruder is or what information he may possess. The privacy measures depend on the types of disclosure~\cite{carvalho2023survey}. Identity disclosure is one of the most critical for data privacy. $k$-anonymity~\cite{samarati2001protecting} is the most popular measure indicating how many individuals share the same information concerning a set of quasi-identifiers, defined according to assumptions on an intruder's background knowledge. A record is unique when $k = 1$, meaning an intruder can single it out. Additionally, linking records between different data sets is an approach that allows measuring the probability of re-identification through different data sets. Record linkage~\cite{fellegi1969theory} is also widely used but focuses on the ability to link records, usually between de-identified data and the original. 

\subsection{Fairness}

Diverse approaches to handling fairness address different parts of the model life-cycle. Several methods to enhance fairness have been proposed in the literature, commonly classified into three categories: pre-processing, in-processing, and post-processing.
We focus on in-processing methods which
involve modifying the machine learning models during training to remove discrimination by incorporating changes into the objective function or imposing a constraint. Adversarial debiasing~\cite{elazar2018adversarial} and exponentiated gradient~\cite{agarwal2018reductions} are prevalent algorithms.

In classification tasks, the most commonly used measures of group fairness include demographic parity~\cite{dwork2012fairness} and equalized odds~\cite{hardt2016equality}. Demographic parity, also known as statistical parity~\cite{dwork2012fairness}, compares the difference in predicted outcome $\hat{Y}$ between any two groups, $|P[\hat{Y} = 1|S = 1] - P[\hat{Y} = 1|S \neq 1]| \leq \epsilon$. Better fairness is achieved with a lower demographic parity value, indicating more similar acceptance rates.
A limitation of this measure is that a highly accurate classifier may be unfair if the proportions of actual positive outcomes vary significantly between groups. Therefore, the equalized odds measure was proposed to overcome such limitation~\cite{hardt2016equality}. This measure computes the difference between the false positive rates $|P[\hat{Y}=1|S=1, Y=0]-P[\hat{Y} = 1|S \neq 1, Y = 0]| \leq \epsilon$, and the difference between the true positive rates of two groups $|P[\hat{Y} = 1|S = 1, Y = 1]-P[\hat{Y} = 1|S \neq 1, Y = 1]| \leq \epsilon$, where smaller differences between groups indicate better fairness.

\subsection{Related Work}\label{subsec:relatedwork}

The increasing interest in synthetic data generation has led to studies on how this type of data protects the individual's privacy and reflects the inherent bias and predictive performance in machine learning applications.

Bhanot et al.~\cite{bhanot2021problem} 
proved the presence of unfairness in generated synthetic data sets and introduced two fairness metrics for time series, emphasizing the importance of evaluating fairness at each evaluation step in the synthetic data generation. Additionally, Chang and Shokri~\cite{chang2021privacy} have shown that fair algorithms tend to memorize data from the under-represented subgroups, increasing the model’s information leakage about unprivileged groups. Their experiments evaluate how and why fair models leak information on synthetic train data.

Machine learning models' efficiency and fairness have also been investigated using synthetic data generated by differentially private GANs. The experiments conducted by Cheng et al.~\cite{cheng2021can} show that integrating differential privacy does not give rise to discrimination during data generation in subsequent classification models. Still, it unfairly amplifies the influence of majority subgroups. Also, the authors demonstrate that differential privacy reduces the quality of the images generated from the GANs and, consequently, the utility in downstream tasks. Recently, Bullwinkel et al.~\cite{bullwinkel2022evaluating} analyzed the interplay between loss of privacy and fairness in the context of models trained on differentially private synthetic data. The experiments focused on binary classification, showing that a notable proportion of the synthesizers studied deteriorated fairness.

The potential of synthetic data in providing privacy-preserving solutions for several data-related challenges and their important role in striving for fairness in machine learning applications prompted our experiments to center around synthetic data generation. 
Although there are exciting and promising approaches for synthetic data generation incorporating differential privacy, the current state of software is still in its early stages, and only DP-CGANS~\cite{sun2022improving} is a viable option for our experiments. However, this tool is considerably time-consuming, and due to this limitation, we do not account for differentially private synthetic data.

Privacy-protected data sets have not yet been analyzed, considering the three vectors of privacy, fairness, and predictive performance. Especially, conclusions about the impact of maximizing each of the vectors remain unclear. Moreover, we focus on the risk of re-identification in privacy-protected data sets rather than membership attacks on predictive models (e.g.~\cite {chang2021privacy}) -- identity disclosure can cause severe consequences for individuals and organizations.

\section{Experimental Study}\label{sec:experiments}
In this section, we provide a thorough experimental study focused on the impact of optimization processes for privacy, fairness, and predictive performance in machine learning.
We aim to answer the following research questions. What are the impacts associated with optimizing a specific vector (\textbf{RQ1}), what are the impacts in prioritizing the remaining vectors (optimization paths) (\textbf{RQ2}), and is there a solution capable of providing a balance between the three vectors (\textbf{RQ3})? 
We describe our experimental methodology in the following sections, briefly describing the data used, methods, and evaluation procedures, followed by presenting experimental results.

\subsection{Data}\label{subsec:data}
In this section, we provide an overview of the commonly used data sets for fairness-aware machine learning~\cite{le2022survey,chakraborty2021bias}. A general description of the main characteristics of these data sets is provided in Table~\ref{tab:data}. The number of attributes and instances were obtained after the cleaning, such as missing data removal. The selection for the protected attributes and quasi-identifiers adheres to previous literature. As fairness measures require protected attributes in a binary form, categorical attributes are grouped; for instance, race=\{caucasian, african-american, hispanic, other\} is transformed to race=\{white, non-white\}, and continuous attributes are discretized like age= \{<25, >=25\}. Such discretization is also defined in the literature and is determined based on privileged and unprivileged groups.

{\setlength\intextsep{2pt}
\begin{table*}[!ht]
\begin{center}
    \scriptsize
    \begin{adjustbox}{max width=\linewidth}

\begin{tabular}{@{}llllll@{}}
\toprule
\textbf{Dataset}             & \textbf{\# Instances} & \textbf{\# Attributes} & \textbf{Domain} & \textbf{Quasi-identifiers}                                                                                                    & \textbf{Protected attributes} \\ \midrule
\textit{Adult}               & 48.842                & 15                     & Finance         & \begin{tabular}[c]{@{}l@{}}Education, age, gender, race, occupation, \\ native country  \end{tabular}                           & Gender, race, age             \\
\textit{German Credit}       & 1.000                 & 22                     & Finance         & \begin{tabular}[c]{@{}l@{}}Purpose, years of employment, age, \\ years of residence, job, gender, foreign worker\end{tabular} & Gender, age                   \\

\textit{Bank marketing}      & 45.211                & 17                     & Finance         & Age, job, marital, education, housing                                                                                         & Age, marital                  \\
\textit{Credit card clients} & 30.000                & 24                     & Finance         & Gender, education, marriage, age                                                                                              & Gender, marriage, education   \\
\textit{COMPAS}              & 6.172                 & 34                     & Criminology     & Gender, age, race, recidivism                                                                                                 & Gender, race                  \\

\textit{Heart disease}       & 1.025                 & 14                     & Healthcare      & Gender, age, heart rate, chest pain                                                                                           & Gender                        \\
\textit{Ricci}               & 118                   & 6                      & Social          & Position, race, combined score                                                                                                & Race                          \\
 \bottomrule
\end{tabular}
\end{adjustbox}  
\caption{General description of the used data sets in the experimental study.}
\label{tab:data}
\end{center}
\end{table*}




\subsection{Methods}

In this section, we describe the \textit{i)} methods used in generating privacy-preserving data variants; \textit{ii)} the learning algorithms and respective hyper-parametrization optimization details employed to generate models, which include standard machine learning (fairness-agnostic) and fairness-aware algorithms; followed by \textit{iii)} evaluation metrics used and \textit{iv)} the overall experimental methodology.

\subsubsection{Synthetic Data Variants}

The synthetic data variants are obtained using two different approaches, PrivateSMOTE and deep learning-based solutions. PrivateSMOTE~\cite{carvalho2022privacy} generates synthetic cases for highest-risk instances (i.e., single-out) based on randomly weighted interpolation of nearest neighbors. We apply PrivateSMOTE with $ratio \in \{1, 2, 3\}$, $knn \in \{1, 3, 5\}$ and $\epsilon \in \{0.1, 0.3, 0.5\}$, where $\epsilon$ is the amount of added noise.
On the other hand, deep learning-based solutions rely on generative models. For comparison purposes, we only synthesize the single-out instances using conditional sampling. Such instances and all attributes are replaced with new cases. 
We leverage the Python SDV package~\cite{sdv} to create different deep-learning variants for this aim. The experiments include Copula GAN, TVAE, and CTGAN with the following parameters: $epochs$ $\in$ $\{100, 200\}$, $batch\_size$ $\in$ $\{50, 100\}$ and \sloppy $embedding\_dim$ $\in$ $\{12, 64\}$. Each set of parameters produces a different synthetic data variant. 

\subsubsection{Learning Algorithms}

There are two types of algorithms used in our experimental evaluation: standard machine learning algorithms (fairness-agnostic) and fairness-aware algorithms. Concerning the former, we leverage three classification algorithms through \textit{Scikit-learn}~\cite{pedregosa2011scikit} toolkit: Random Forest~\cite{ho1998random}, XGBoost~\cite{chen2016xgboost} and Logistic Regression~\cite{logit}. Final models for each algorithm are chosen based on a 2*5-fold cross-validation estimation of evaluation scores for models based on a grid search method. For fairness mitigation, we use FairMask~\cite{peng2022fairmask} and exponentiated gradient from Fairlearn~\cite{bird2020fairlearn}. Table~\ref{tab:algorithms} summarizes this information.

{\setlength\intextsep{2pt}
\begin{table}[!ht]
\begin{center}
    \scriptsize
    \begin{adjustbox}{max width=0.4\linewidth}
\begin{tabular}{@{}l|l@{}}
\toprule
\textbf{Algorithm}  & \textbf{Parameters}                                       \\ \midrule
Random Forest       & \begin{tabular}[c]{@{}l@{}}$n\_estimators \in \{100, 250, 500\}$\\ $max\_depth \in \{4, 7, 10\}$\end{tabular}                                                                                       \\ \midrule
Boosting            & \begin{tabular}[c]{@{}l@{}}$n\_estimators \in \{100, 250, 500\}$\\ $max\_depth \in \{4, 7, 10\}$\\ $learning\_rate \in \{0.1, 0.01\}$\end{tabular}                                                                     \\ \midrule
Logistic Regression & \begin{tabular}[c]{@{}l@{}}$C \in \{0.001, 1, 10000\}$\\ $max\_iter \in \{10e^5, 10e^6\}$\end{tabular}                                                                                      \\ \bottomrule
\end{tabular}
\end{adjustbox}
\end{center}    
\caption{Learning algorithms and respective hyper-parameter grid used in the experimental study.}
    \label{tab:algorithms}
\end{table}

\subsubsection{Evaluation}\label{subsubsec:methods}

All synthetic data variants are evaluated in terms of re-identification risk, fairness, and predictive performance. 

To assess the potential of re-identification, we use the \textit{Python Record Linkage Toolkit}~\cite{de_bruin_j_2019_3559043} to compare each variant with the original considering a specified set of quasi-identifiers. In this study, we focus on exact matches, where all values for the quasi-identifiers match, resulting in a 100\% likelihood of re-identification. Such comparisons are carried out in the sets of single-out instances. 
In the learning phase, we use equalized odds difference for fairness evaluation and Accuracy for predictive performance concerning the testing data. 

\subsubsection{Experimental Methodology}\label{subsubsec:methods}
For conciseness, our experimental methodology is illustrated in Figure~\ref{fig:flux}.
The experimental study begins by splitting each original data into training and test sets corresponding to 80\% and 20\%, respectively. Then, we generate several synthetic data variants using the training data set for privacy constraints, in which re-identification risk is evaluated by comparing each synthetic data variant to the original data. Then, models are generated using both fairness-agnostic and fairness-aware algorithms. After the training phase, out-of-sample predictive performance and fairness of the models are measured. 

{\setlength\intextsep{2pt}
\begin{figure*}[ht!]
   \centering
   \includegraphics[width=0.9\linewidth]{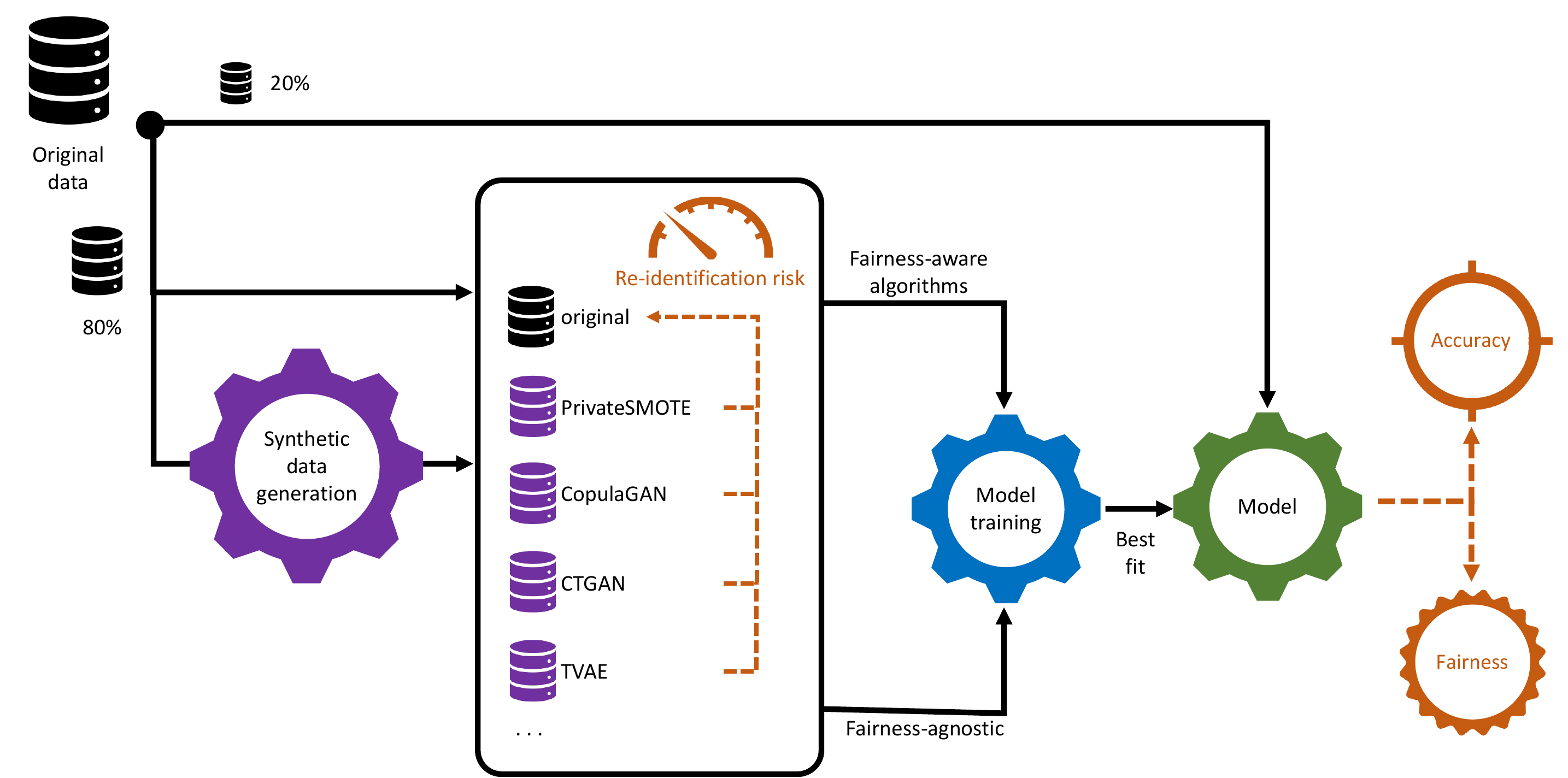}
 \caption{Workflow of our experimental methodology.}
 \label{fig:flux}
\end{figure*}

\subsection{Experimental Results}\label{subsec:results}
The following set of results refers to the probability of each optimized vector winning or losing when compared to the remaining vectors. We construct optimization paths, as demonstrated in Figure~\ref{fig:paths}, to analyze the relevance of prioritizing a specific vector.
To calculate the probabilities, we select the best models estimated via cross-validation in out-of-sample. Each solution, i.e., privacy-protected data variant outcome, is compared to a baseline.
For visual purposes, "A@", "FM@" and "FL@" refers to the fairness-agnostic and fairness-aware algorithms, namely, Agnostic, FairMask, and Fairlearn. 

\begin{figure*}[ht!]
   \centering
   \includegraphics[width=0.7\linewidth]{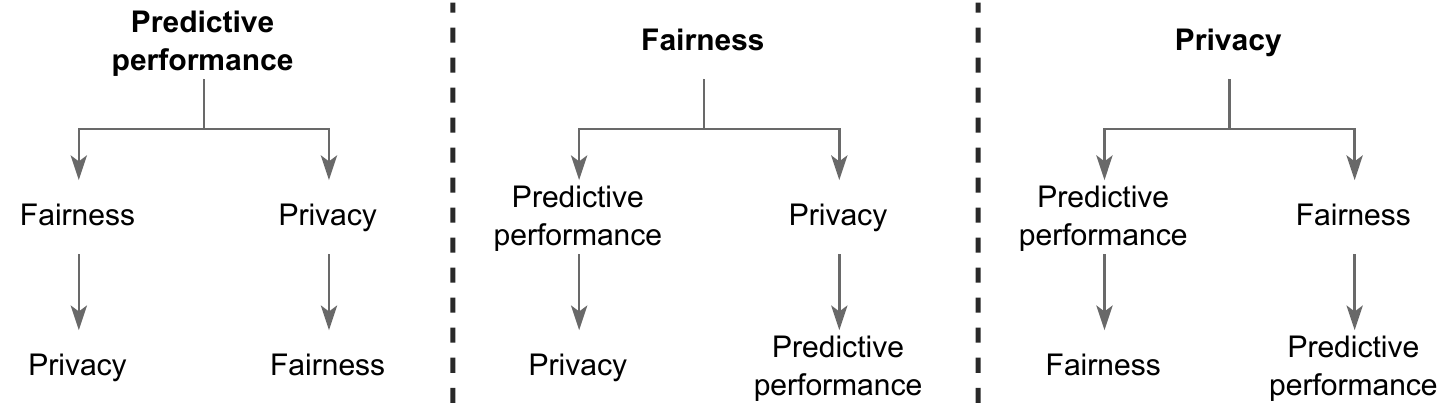}
 \caption{Optimization paths for each vector.}
 \label{fig:paths}
\end{figure*}

\paragraph{Predictive performance vector.} The left image in Figure~\ref{fig:acc} shows the probabilities of each solution's fairness winning and losing against the baseline as well as the probabilities of privacy winning and losing, while the right image shows the reverse path. Such a baseline corresponds to the best solution in terms of predictive performance for each data set. Note that the probabilities refer to the percentage of the cases in the total number of privacy-protected data variants for each fairness-agnostic and fairness-aware algorithms. Results show that models optimized towards predictive performance demonstrate a balanced probability of winning or losing w.r.t. fairness. Also, when the models outperform the baseline, the solutions tend to be more private, however, the opposite is not necessarily true: the reverse path shows that except for PrivateSMOTE, all synthetic approaches outperform the baseline in terms of privacy. Therefore, optimizing predictive performance results in losses for privacy but it is possible to attain fairer models to a certain extent.
Additionally, we observe that Fairlearn leads to fairer and more private solutions.

{\setlength\intextsep{2pt}
 \begin{figure*}
\begin{subfigure}[h]{0.5\linewidth}
\includegraphics[width=\linewidth]{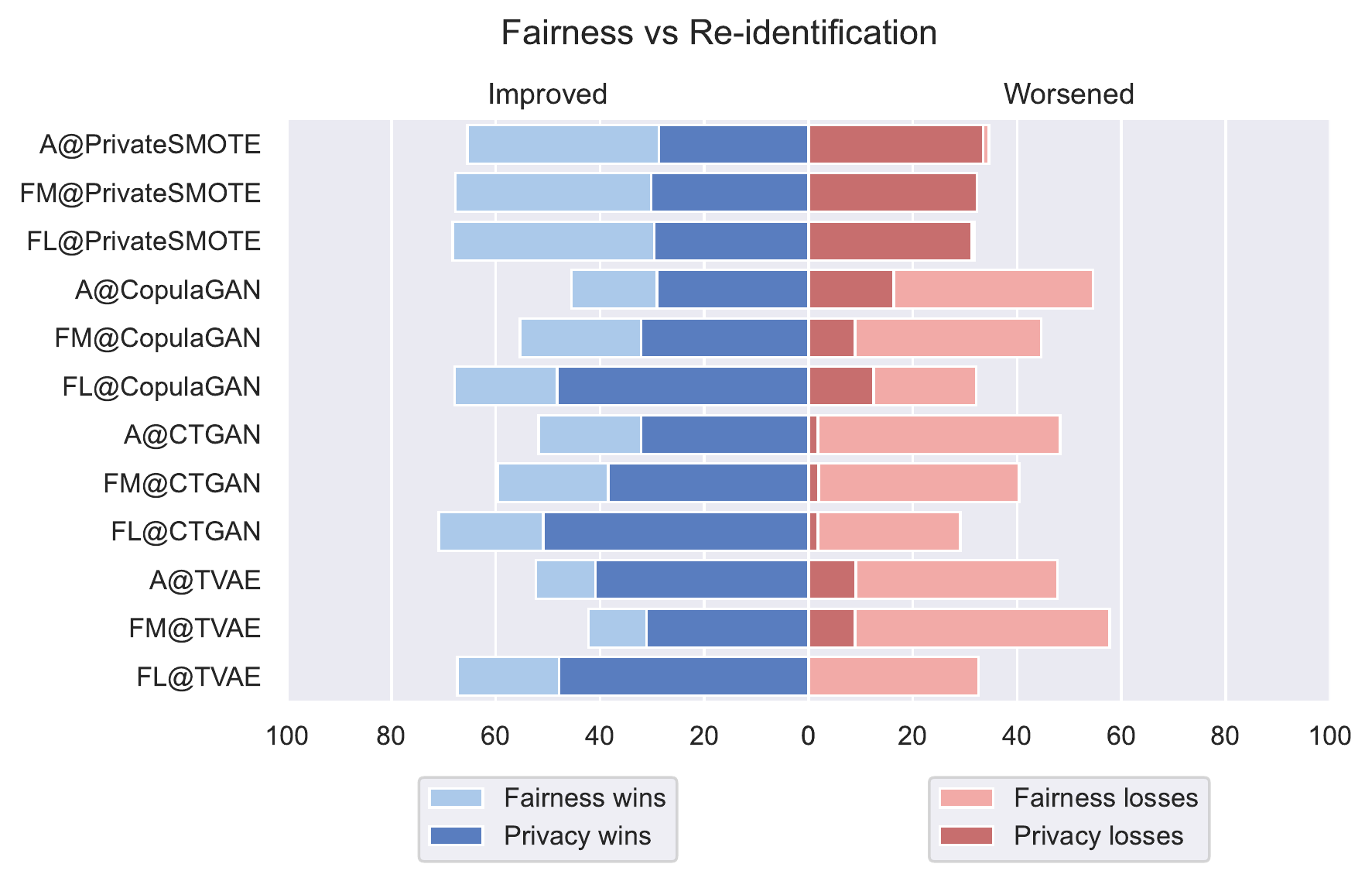}
\end{subfigure}
\hfill
\begin{subfigure}[h]{0.5\linewidth}
\includegraphics[width=\linewidth]{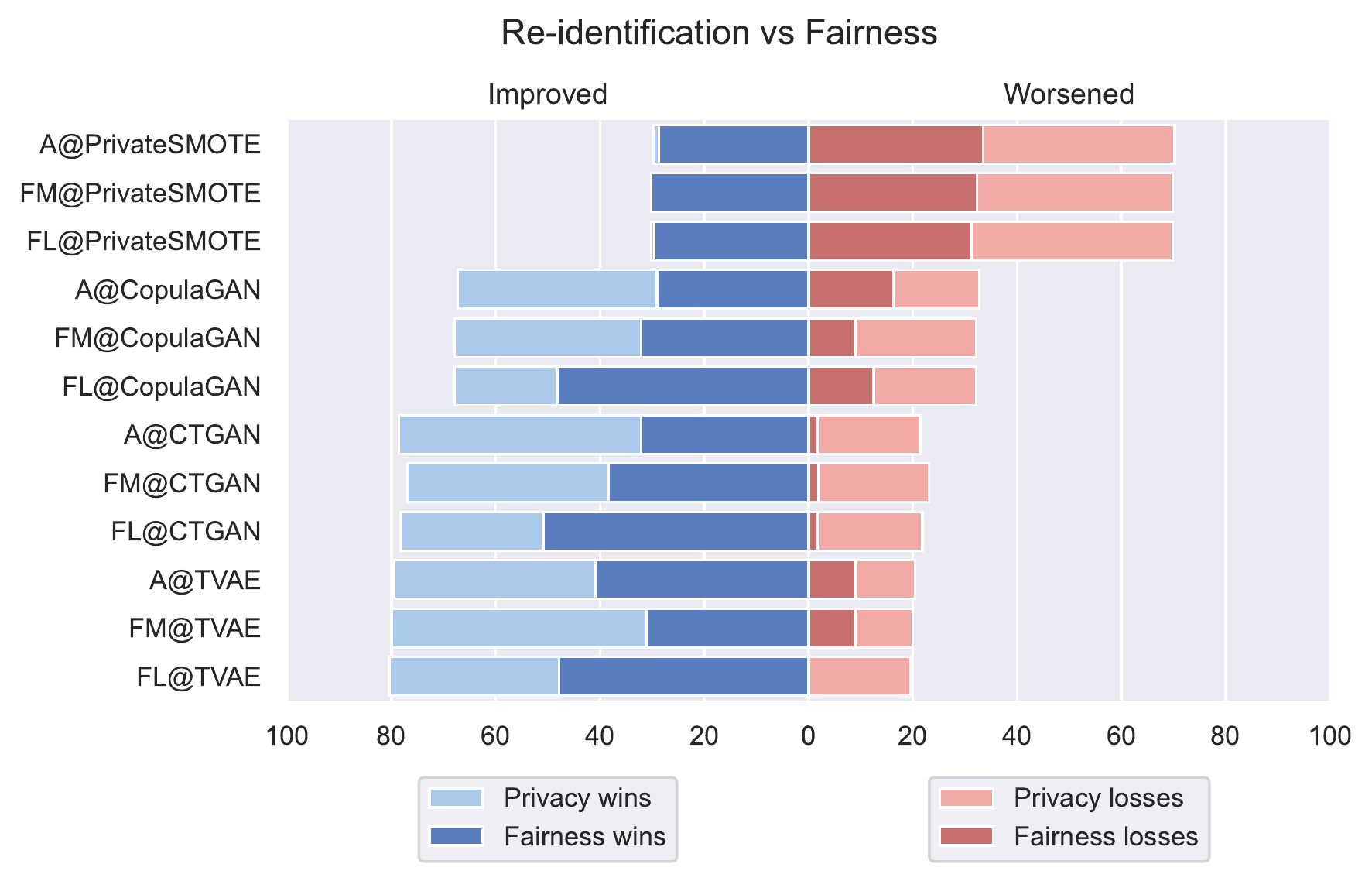}
\end{subfigure}%
\caption{Predictive performance optimization paths. Total wins/losses comparing each solution to the baseline (best solution in terms of Accuracy) for fairness along with the respective wins for privacy (left) and vice-versa (right).}
\label{fig:acc}
\end{figure*}

\paragraph{Fairness vector.} Concerning this vector, Figure~\ref{fig:fairness} illustrates the probabilities of each solution's predictive performance winning or losing compared to the baseline (best solution in terms of equalized odds) with the respective probabilities of privacy winning and vice-versa. A notable outcome is the outperformed solutions in terms of predictive performance with the same probability of privacy wins. In the reverse path, the majority of solutions present a lower re-identification risk compared to the baseline. Besides, the models for such solutions present a probability equal to or higher than 50\% of improving predictive performance.  

{\setlength\intextsep{2pt}
\begin{figure*}
\begin{subfigure}[h]{0.5\linewidth}
\includegraphics[width=\linewidth]{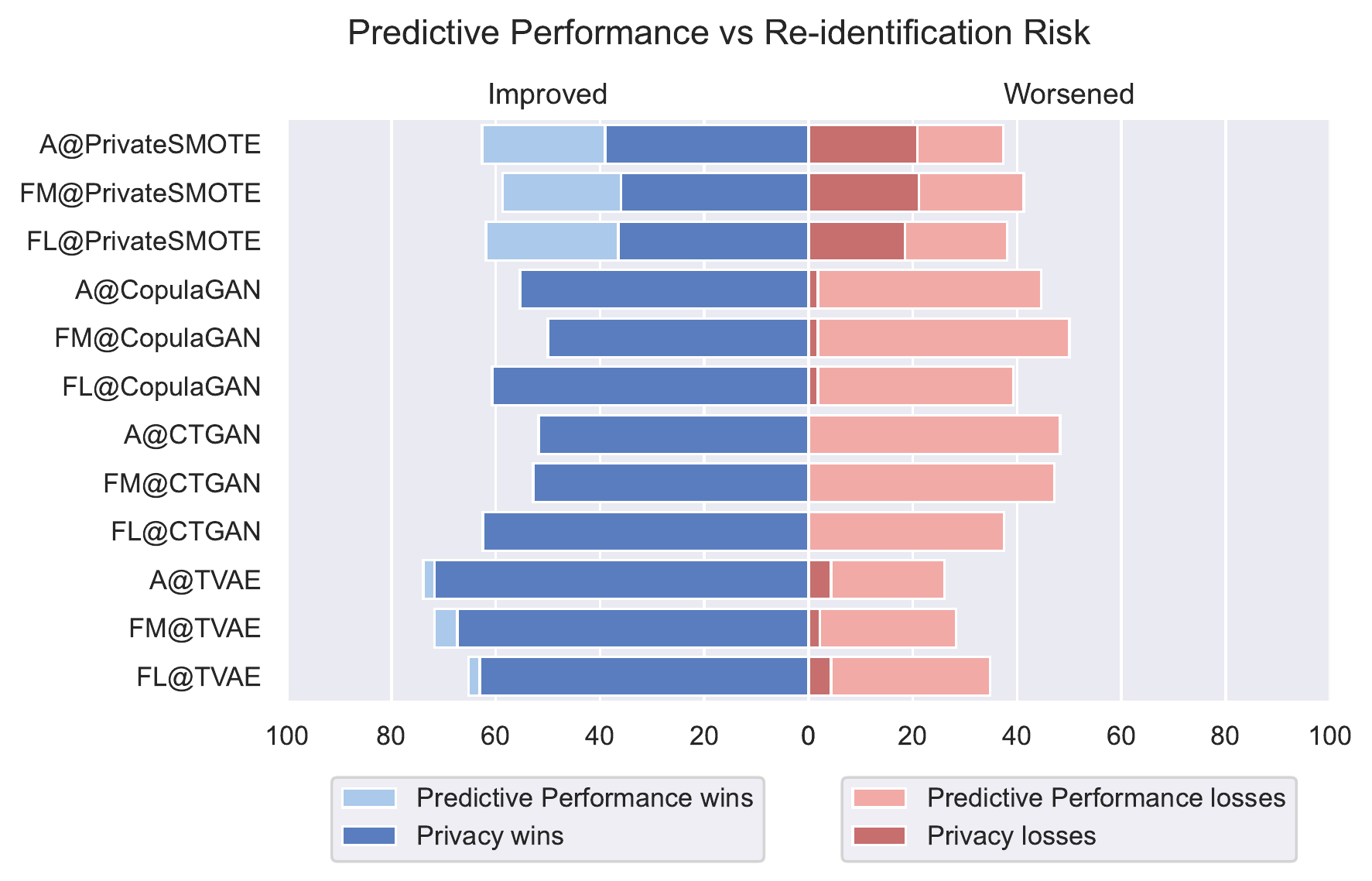}
\end{subfigure}
\hfill
\begin{subfigure}[h]{0.5\linewidth}
\includegraphics[width=\linewidth]{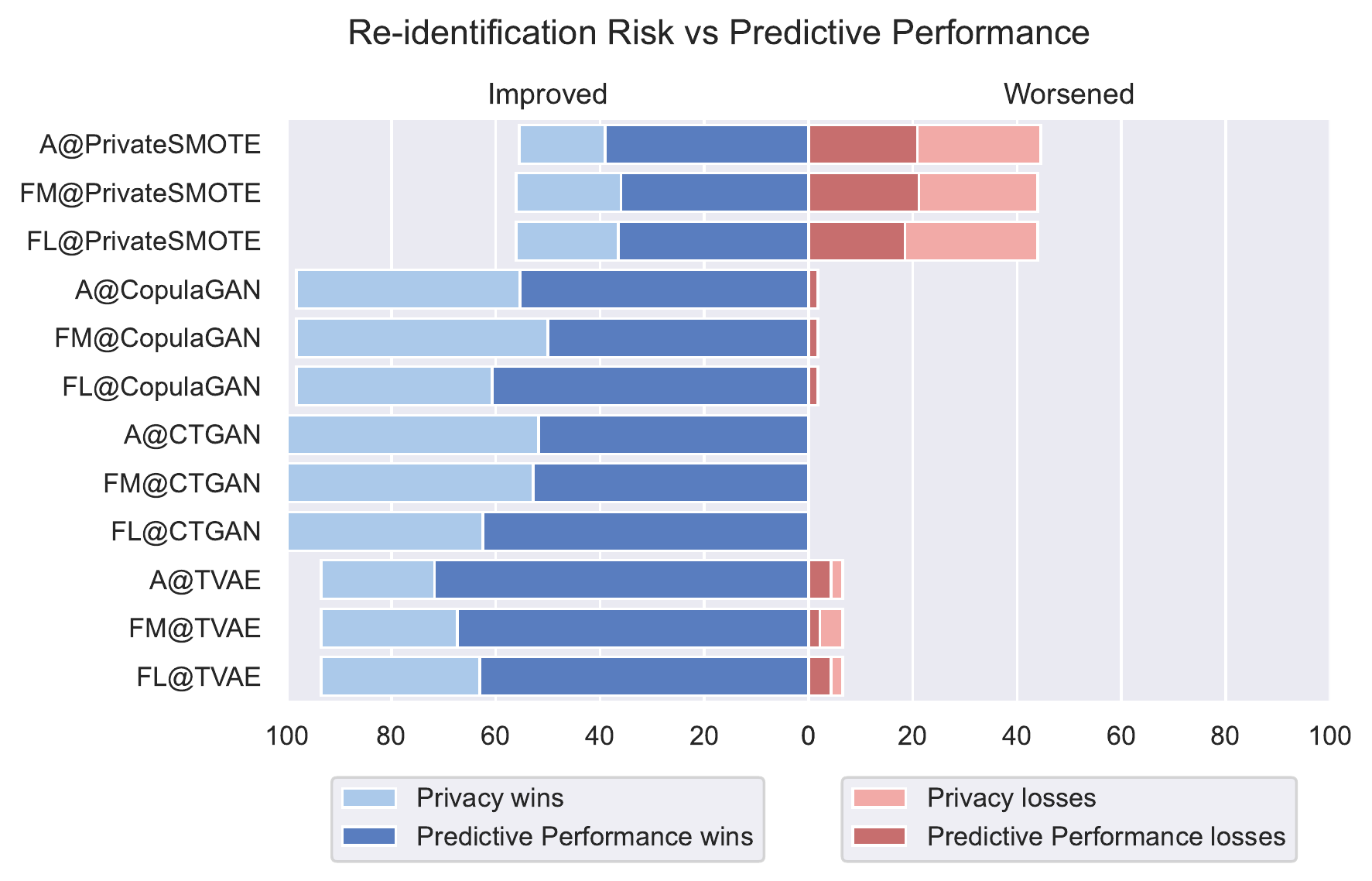}
\end{subfigure}%
\caption{Fairness optimization paths. Total wins/losses comparing each solution to the baseline (best solution in terms of equalized odds) for predictive performance along with the respective wins for privacy (left) and vice-versa (right).}
\label{fig:fairness}
\end{figure*}

\paragraph{Privacy vector.} Lastly, Figure~\ref{fig:privacy} illustrates the total wins and losses of each solution's predictive performance compared to the baseline (best solution in terms of re-identification risk) along with the respective probabilities of fairness winning and vice-versa. In this scenario, the baseline has a probability greater than 50\% of outperforming the remaining solutions in w.r.t. predictive performance. 
On the other hand, when we prioritize fairness, the baseline tends to lose. 

{\setlength\intextsep{2pt}
\begin{figure*}
\begin{subfigure}[h]{0.5\linewidth}
\includegraphics[width=\linewidth]{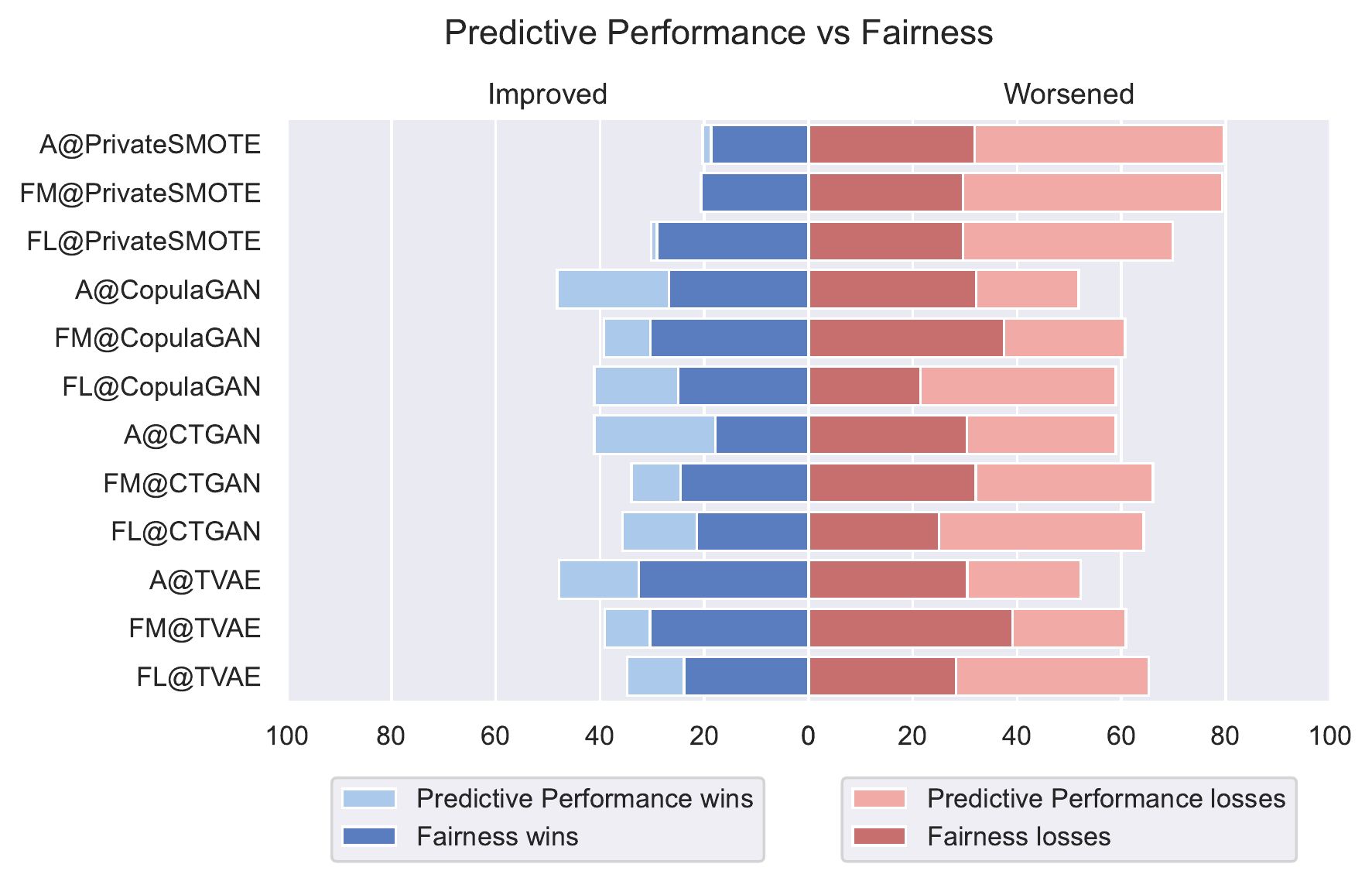}
\end{subfigure}
\hfill
\begin{subfigure}[h]{0.5\linewidth}
\includegraphics[width=\linewidth]{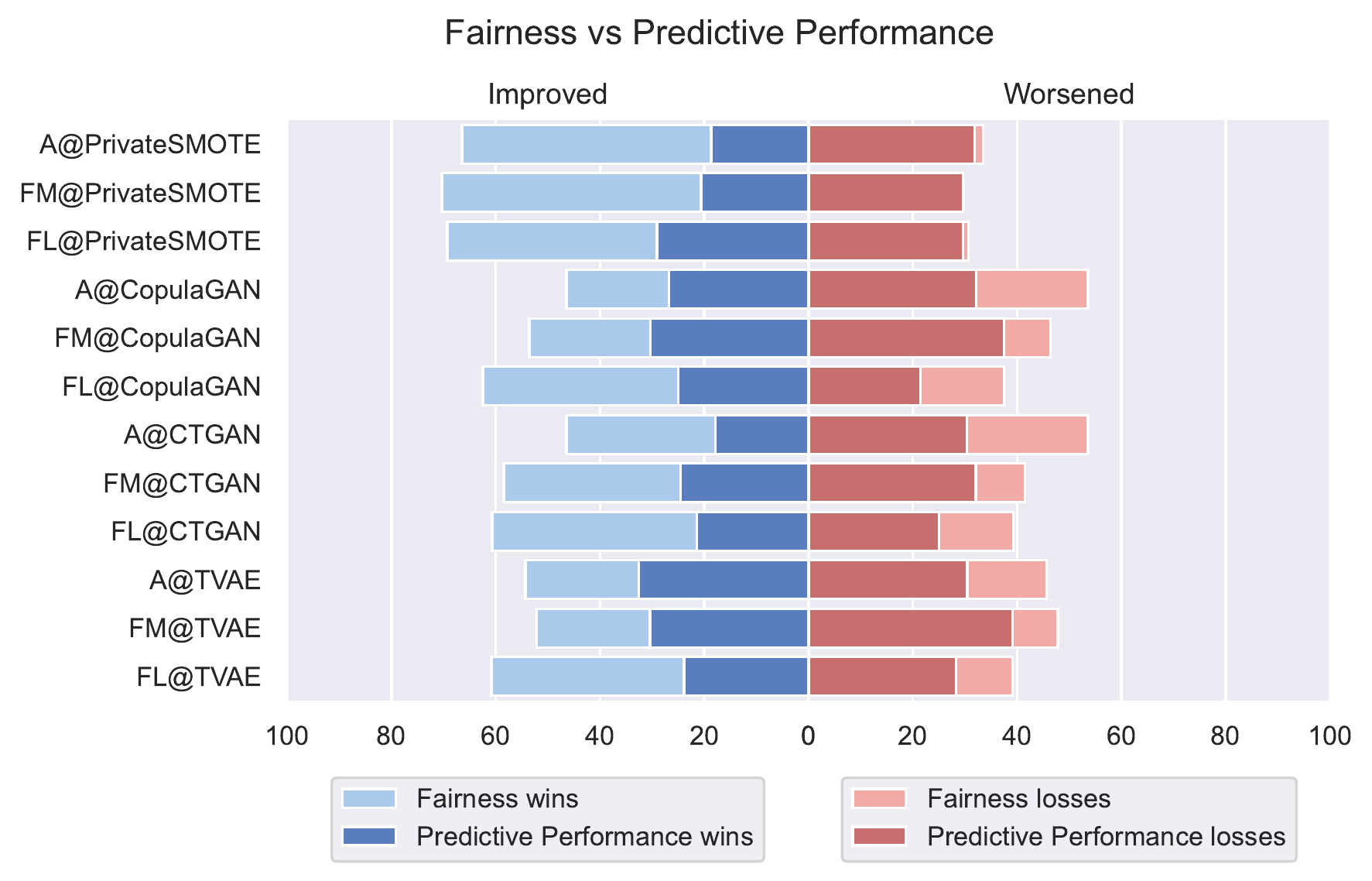}
\end{subfigure}%
\caption{Privacy optimization paths. Total wins/losses comparing each solution to the baseline (best solution in terms of re-identification risk) for predictive performance, and respective wins for fairness (left) and vice-versa (right).}
\label{fig:privacy}
\end{figure*}

\paragraph{All vectors optimized.} In the previous set of results, we show the impacts of optimizing a single vector. However, an optimal solution should maintain a balance across all the vectors. Therefore, we aim to analyze to what extent is possible to obtain such a balance.
Figure~\ref{fig:bayes} provides a comparison reporting the statistical tests using the Bayes Sign Test~\cite{Benavoli2014,Benavoli2017} with a ROPE interval of [-1\%, 1\%] to evaluate the statistical significance concerning the percentage difference for each vector optimization. This percentage is defined as $\frac{R_a - R_b}{R_b} * 100$ where $R_a$ is the solution under comparison and $R_b$ is the baseline. ROPE (Region of Practical Equivalence)~\cite{kruschke2015bayesian} is used to specify the probability of the difference of values being inside a specific interval as having practically no effect. If the percentage difference is within the specified range, they are of practical equivalence (draw), and if the percentage difference is less than -1\%, \textit{b} outperforms solution \textit{a} (lose). Such a baseline corresponds to the best solutions for each vector while the solutions under comparison correspond to the ones with the best average rank across the three vectors for each data set.

Figure~\ref{fig:bayes} shows the comparisons for each synthetic approach between the optimal solutions for each vector and the solutions with the best-averaged rank across all vectors. Concerning predictive performance, the average rank solutions' models are, for the most part, capable of providing practical equivalence to the optimal solutions of this vector with a probability higher than 50\%. 
Although the privacy vector presents higher losses for some solutions, A@CopulaGAN and A@CTGAN stand out with a probability of drawing to the baseline greater than 80\%. However, the models for these solutions are less fair. Additionally, such an outcome shows that it may be possible to obtain a balance between the three vectors, through TVAE-based solutions.

\begin{figure*}
\centering
\includegraphics[width=.57\linewidth]{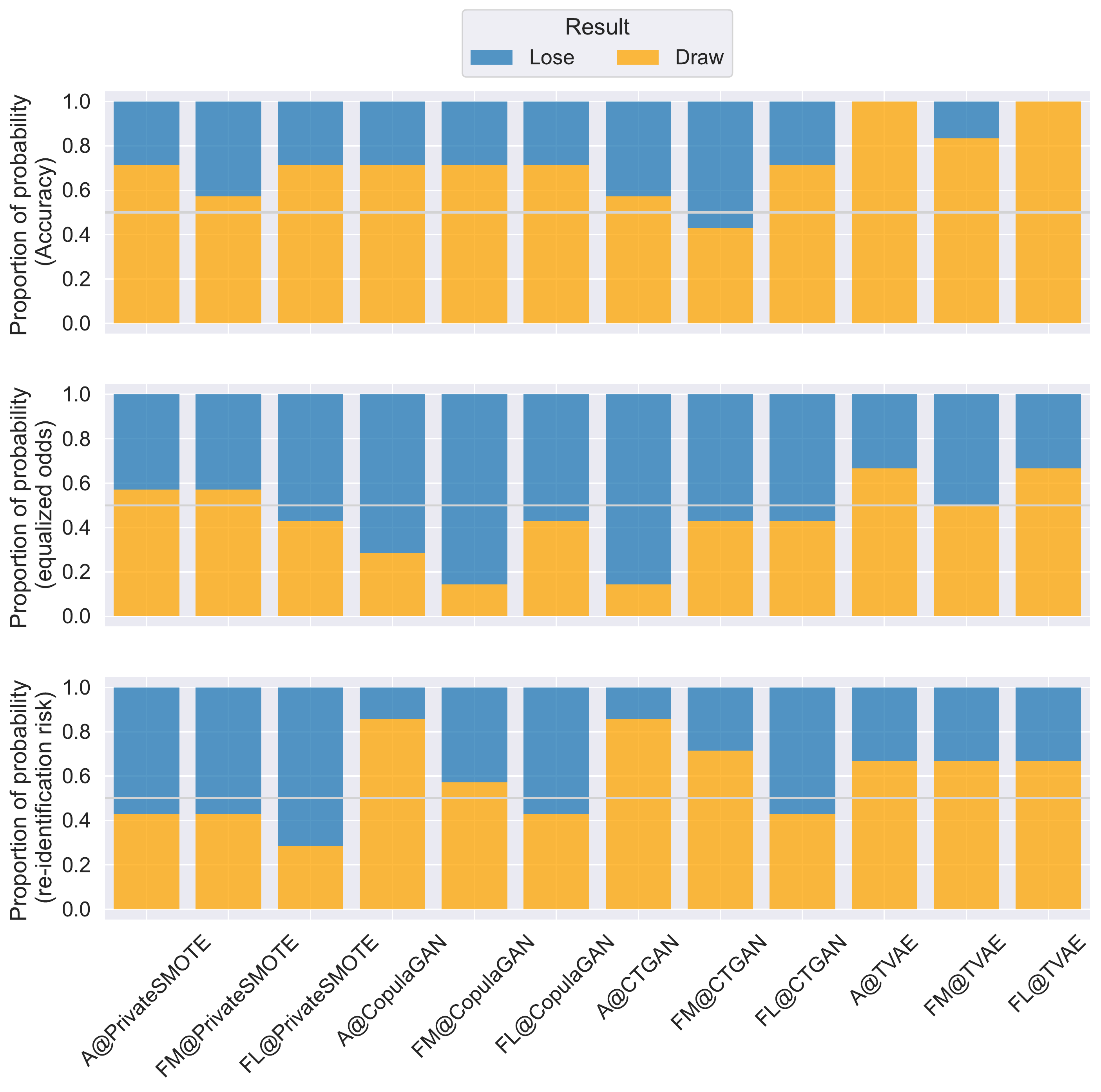}
\caption{Proportion of probability for each candidate
solution drawing or losing significantly against the solution with the best-averaged rank between the three vectors, according to the Bayes Sign Test.}
\label{fig:bayes}
\end{figure*}

\section{Discussion}\label{sec:discussion}

Given the results of the experimental evaluation presented above, conclusions imply that developing safe machine learning applications -- one that protects individual data privacy and prevents disparate treatment with a negative impact on protected groups, may face even greater challenges than currently construed. This leads to questions that require attention in future work:

\begin{itemize}
    \item \textbf{(RQ1)} A notable finding for future work is the relationship between the different optimization vectors. Figure~\ref{fig:acc} and~\ref{fig:fairness}, show that many solutions tested and which do not improve their respective optimization vector exhibit a considerable ability to reduce re-identification risk when privacy is the priority in the optimization path. 
    Also, we observe in Figures~\ref{fig:acc} and~\ref{fig:privacy} that optimizing predictive performance or privacy shows a similar impact on fairness.
  
    \item \textbf{(RQ2)} Although it is unlikely to obtain a solution with practically no losses, the optimization paths allow us to find which vector should be prioritized to prevent higher losses. When optimizing privacy (Figure~\ref{fig:privacy}), and prioritizing predictive performance, we observe that the majority of the solutions maintain the ability to produce accurate models. Such an outcome shows that it is possible to obtain private solutions with minimal impact on predictive performance but this comes at the expense of fairness.

    \item \textbf{(RQ3)} Nevertheless, finding a solution that balances the three vectors is crucial. However, as shown in Figure~\ref{fig:bayes}, it is, in general, very improbable to achieve a good balance between all vectors. Despite our experiments demonstrating that TVAE-based solutions are, to a certain degree, capable of obtaining such a balance, that does not happen for the remaining approaches.
    
\end{itemize}

\section{Conclusion}\label{sec:conclusion}


This paper thoroughly analyzes the dynamics between privacy, fairness, and predictive performance by assessing the impact of optimizing a single vector on the remaining vectors. We generate multiple privacy-protected data variants from the original data using synthetization methods and evaluate each variant in terms of privacy w.r.t re-identification risk but also fairness and predictive performance for both fairness-agnostic and fairness-aware algorithms. 

The main conclusions indicate that in single vector optimization, the remaining vectors will suffer from losses. Nevertheless, optimizing privacy and prioritizing predictive performance allows for obtaining private solutions while maintaining the predictive performance intact. However, it is difficult to navigate a balance between the three vectors. These results highlight the importance of further developments in discriminatory bias when the goal is to release or share personal information. For future work, we plan to analyze the effects of data preparation on fairness as the presence of inherent biases in a data set may pose challenges in achieving fairer models~\cite{valentim2019impact}.
The Python code and data necessary to replicate the results shown in this paper are available at \textit{https://tinyurl.com/yku3s7du}.


%
%
%
\bibliographystyle{splncs04}
\bibliography{bibs}
%




\end{document}